%% file: main.tex
\documentclass[10pt,twocolumn,letterpaper]{article}
\usepackage{template}
\definecolor{myblue}{rgb}{0.21,0.49,0.74}
\usepackage[pagebackref,breaklinks,colorlinks,allcolors=myblue]{hyperref}
\usepackage{makecell}
\usepackage{xcolor}
\usepackage{float}

\title{SDAKD: Student Discriminator Assisted Knowledge Distillation for Super-Resolution Generative Adversarial Networks}

\author{Nikolaos Kaparinos\\
CERTH-ITI\\
Thessaloniki, Greece, 57001\\
{\tt\small kaparinos@iti.gr}
\and
Vasileios Mezaris\\
CERTH-ITI\\
Thessaloniki, Greece, 57001\\
{\tt\small bmezaris@iti.gr}
}

\begin{document}
\maketitle
\input{sec/0_abstract}    
\input{sec/1_intro}

\input{sec/2_related_work}
\input{sec/3_proposed_method}
\input{sec/4_experiments}
\input{sec/5_conclusions}

\end{document}

%% file: sec/0_abstract.tex
\begin{abstract}
Generative Adversarial Networks (GANs) achieve excellent performance in generative tasks, such as image super-resolution, but their computational requirements make difficult their deployment on resource-constrained devices. While knowledge distillation is a promising research direction for GAN compression, effectively training a smaller student generator is challenging due to the capacity mismatch between the student generator and the teacher discriminator. In this work, we propose Student Discriminator Assisted Knowledge Distillation (SDAKD), a novel GAN distillation methodology that introduces a student discriminator to mitigate this capacity mismatch. SDAKD follows a three-stage training strategy,
and integrates an adapted feature map distillation approach in its last two training stages. We evaluated SDAKD on two well-performing super-resolution GANs, GCFSR and Real-ESRGAN. Our experiments demonstrate consistent improvements over the baselines and SOTA GAN knowledge distillation methods. 
The SDAKD source code will be made openly available upon acceptance of the paper.
\end{abstract}

%% file: sec/1_intro.tex
\section{Introduction}
\label{sec:intro}
Generative Adversarial Networks (GANs) are a prominent class of generative neural network models, capable of producing highly realistic images, videos and other complex data distributions. They have been successfully applied across diverse tasks, including but not limited to super-resolution \cite{he2022gcfsr, ledig2017photo}, image generation \cite{karras2019style, cao2018recent}, natural language generation \cite{zhang2016generating, guo2018long} and data augmentation for downstream tasks \cite{xie2020unsupervised, frid2018synthetic, antoniou2018augmenting, shin2018medical}. Even with the emergence of diffusion models, GANs remain relevant, particularly since diffusion models suffer from slow output generation: they need multiple forward passes to generate a single output, whereas GANs require a single pass through the generator.

Despite this advantage, typical state-of-the-art (SOTA) GAN architectures still demand substantial memory and computation resources, making difficult their use in resource-constrained environments such as mobile devices, autonomous systems and embedded platforms. One class of techniques for alleviating this difficulty is neural architecture search (NAS), \ie searching for compact GAN architectures without human engineering, e.g., \cite{fu2020autogan, jin2021teachers, lu2025growing, xue2025architecture}. While NAS can yield efficient models, it often requires excessive computational resources during the search phase. A more cost-effective  class of techniques is knowledge distillation, \ie transferring knowledge from a high-capacity teacher model to a smaller and more efficient student model. However, applying knowledge distillation to GANs is particularly challenging because adversarial training requires balancing the generator`s and discriminator`s capacity. 

In this work, we propose a GAN knowledge distillation methodology that introduces a second discriminator, i.e., a student discriminator, to mitigate the capacity mismatch between the student generator and the teacher discriminator. This second discriminator contributes to more effective adversarial training of the student generator. We test our proposed approach on two super-resolution GAN models, GCFSR \cite{he2022gcfsr} and Real-ESRGAN \cite{wang2021real}. 

Specifically, our main contributions are the following:
\begin{itemize}
\item We propose the use of a student discriminator, whose initial knowledge is distilled from the pre-trained teacher discriminator, to mitigate the capacity mismatch between the student generator and the teacher discriminator.
\item We introduce a new stage in the typical distillation pipeline, where the two student networks are first trained only via supervised learning, prior to the training stage that combines supervised and adversarial training.
\item To introduce feature map distillation in the proposed distillation pipeline, we adapt the MLP \cite{liu2023simple} method to accommodate student network architectures with fewer channels than their teacher counterparts.
\end{itemize}

%% file: sec/2_related_work.tex
\section{Related Work}
\subsection{Generative Adversarial Networks}
Generative Adversarial Networks (GANs), introduced by Goodfellow \etal \cite{goodfellow2020generative}, are a class of generative neural network models that frame the learning procedure as a two-player minimax game. The players in this minimax game are two neural networks, the generator and the discriminator. The objective of the generator is to learn the underlying data distribution of the given dataset and to be able to generate new samples from this distribution. In contrast, the discriminator is a binary classifier whose objective is to distinguish between real and fake generated samples. Hence, the output of the discriminator is the probability that the input is real, denoted as $P_{real}$. GANs have demonstrated notable success in various tasks, including but not limited to image generation \cite{radford2015unsupervised, cao2018recent}, style transfer \cite{karras2019style}, super-resolution \cite{wang2018esrgan, he2022gcfsr, wang2021real},  text-to-image generation \cite{xu2018attngan}, natural language generation \cite{zhang2016generating, guo2018long} and data augmentation \cite{xie2020unsupervised, frid2018synthetic, antoniou2018augmenting, shin2018medical}. The success of GANs has led to an exponential increase in related research \cite{cheng2020generative}. DCGAN \cite{radford2015unsupervised} introduced deep convolutional architectures into GANs. Two notable convolutional GAN models are Pix2Pix and StyleGAN; Pix2Pix \cite{isola2017image} utilized a GAN model for the task of image-to-image translation, while StyleGAN \cite{karras2019style} enabled control in image generation. SRGAN \cite{ledig2017photo} was the first GAN-based model proposed for the image super-resolution task. Additionally, SRGAN introduced the perceptual loss function to enhance super-resolution performance. ESRGAN \cite{wang2018esrgan} tuned the architecture and the loss function of SRGAN to further increase performance. Real-ESRGAN \cite{wang2021real} further built upon ESRGAN by employing a degradation modeling process to improve performance in real-world degraded images. GCFSR \cite{he2022gcfsr} achieved SOTA results in face super-resolution without facial or GAN priors by employing an encoder-generator architecture that utilizes style and feature modulation.

\subsection{Knowledge Distillation Overview}
Knowledge distillation is a broad class of model compression techniques where a smaller and computationally efficient student model is trained to replicate a larger teacher model. By transferring knowledge from the teacher model, knowledge distillation enables smaller models to achieve performance similar to that of larger ones, while having lower computational and memory requirements \cite{gou2021knowledge}. 

In response-based knowledge distillation, the student network is trained to mimic the response of the last output layer of the teacher \cite{moslemi2024survey}. Despite its simplicity, response-based knowledge distillation is an effective strategy that is widely applied in various tasks. Feature Map (FM) based knowledge distillation, on the other hand, aims to transfer knowledge about intermediate feature representations generated by the teacher. This is a valuable addition to response-based distillation, as intermediate representations can provide rich semantic information \cite{gou2021knowledge}.

The concept of feature-based knowledge distillation was originally introduced by FitNet \cite{romero2014fitnets}, which proposed using a trainable regressor to transform the student’s FMs to match those of the teacher and subsequently training the student network. One limitation of FitNet, however, is that the FM transformation network is trained separately from the student network, resulting in increased overall training time. Additionally, FitNet was designed specifically for image classification, limiting its applicability to other tasks. KR \cite{chen2021distilling} utilized cross-stage connection paths to distill knowledge from multiple teacher layers to a single student FM. However, one drawback of KR is its computational complexity, as the hierarchical cross-stage connections increase the overall processing overhead. MGD \cite{yang2022masked} introduced a generative strategy in which parts of the student’s FMs are randomly masked and an adversarial generator is used to encourage the student to reconstruct features that resemble those of the teacher. CWD \cite{shu2021channel} minimized the KL divergence between normalized channel activation maps. 
MLP \cite{liu2023simple} utilized a small trainable network to transform the FM from the student’s backbone final convolutional layer and then applied a L2 loss between the transformed output and the corresponding teacher`s FM. MLP consistently achieves SOTA performance or is on par with the SOTA across various networks and tasks, 
and is computationally efficient and straightforward to implement. Although the MLP approach was originally proposed for supervised learning tasks, such as image classification, object detection, instance segmentation and semantic segmentation, we chose to apply it in the context of GANs due to its demonstrated benefits.

\begin{figure*}[]
    \centering
    \includegraphics[width=1.00\linewidth]{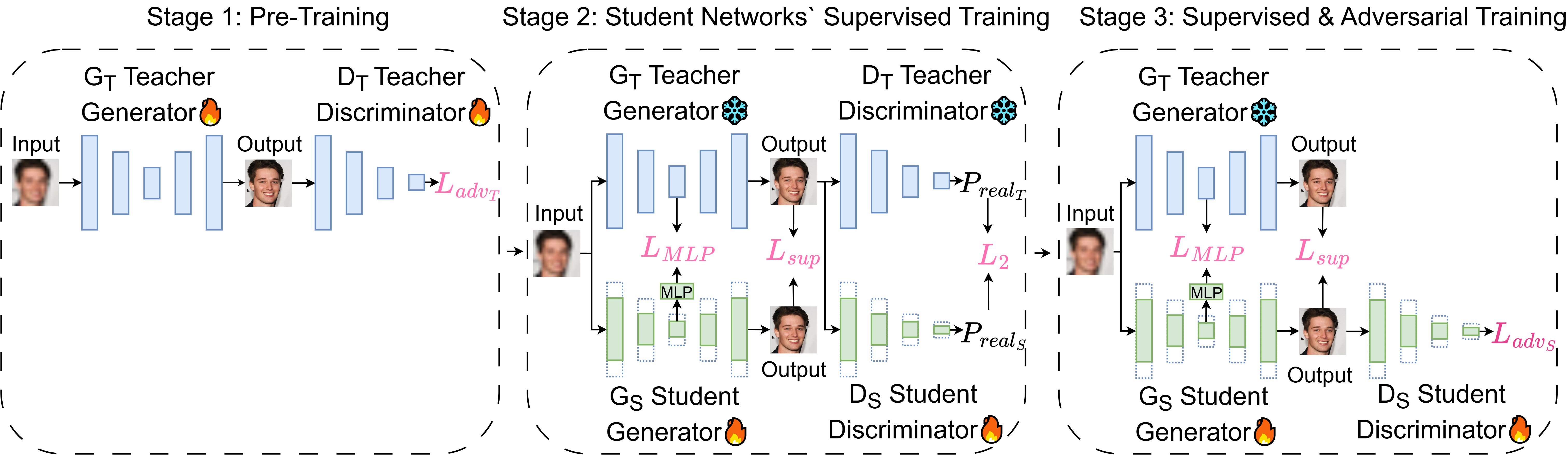}
    \caption{Overview of SDAKD, our proposed GAN knowledge distillation methodology.}
    \label{fig:main}
\end{figure*}

\subsection{Knowledge Distillation for GANs}
\label{kd-in-gans}
Generic model compression strategies could be applied to GANs, however they are not optimal, as they do account for their unique characteristics. For this, various GAN-specific model compression techniques have been proposed. Co-Evolution \cite{shu2019co} proposes an evolutionary channel pruning algorithm, representing the first model compression technique specifically designed for GAN models. GAN Slimming \cite{wang2020gan} proposes a compression algorithm that includes pruning, knowledge distillation and quantization. GAN-Compression \cite{li2020gan} combines NAS and knowledge distillation by first searching for an efficient student generator architecture and, subsequently, supervising its training with both the teacher generator and discriminator. OMGD \cite{ren2021online} utilizes knowledge distillation to train a discriminator-free student generator that is supervised by two teacher generators. DMAD \cite{li2022learning} leverages differentiable masks to progressively identify a lightweight student generator architecture while training it under the supervision of the teacher networks with both response and FM distillation. Wavelet KD \cite{zhang2022wavelet} decomposes the images into frequency bands and distills only the high-frequency ones, which are typically more challenging for GAN student models to learn. Furthermore, task-specific GAN knowledge distillation techniques have also been proposed, \eg, for the tasks of face-swapping \cite{ding2025smfswap} and underwater image enhancement \cite{zhang2025zrlgans}.

Applying knowledge distillation to GANs presents unique challenges compared to supervised learning scenarios because of the presence of two networks, the generator and the discriminator. For effective adversarial training, both networks must have comparable capacity. Otherwise, issues such as mode collapse arise during training \cite{iglesias2023survey, chakraborty2024ten}. Additionally, previous studies have shown that training a high performing student generator requires the supervision of both the teacher generator and the teacher discriminator \cite{li2020gan, hu2023discriminator}. However, this is challenging as the teacher discriminator network is designed to approximately match the capacity of the teacher generator, which is larger than the student generator`s capacity. Hence, there exists a capacity mismatch between the teacher discriminator and the student generator.
Prior work in the literature has explored various strategies to mitigate the capacity mismatch and ensure effective and stable adversarial training of the student generator. GCC \cite{li2021revisiting} selectively activates only a subset of the discriminator`s neurons during the student generator`s adversarial training. DCD \cite{hu2023discriminator} employs online knowledge distillation, in which the teacher generator, teacher discriminator and the student generator are all trained simultaneously from scratch. By avoiding a pre-trained discriminator, this approach partially mitigates the capacity mismatch, achieving SOTA performance in GAN distillation.

%% file: sec/3_proposed_method.tex
\section{Proposed Methodology}
\subsection{Overall Approach}
\label{over-appr}

In this study, we address the challenge of compressing a GAN model by first training the teacher models (teacher generator $G_T$ and teacher discriminator $D_T$) and then applying knowledge distillation to train a student generator $G_S$. The student generator $G_S$ is a smaller and more efficient network compared to its teacher counterpart, since it has a reduced number of channels in its convolutional layers. Specifically, similarly to previous works in the GAN knowledge distillation literature \cite{ren2021online, li2021revisiting, hu2023discriminator}, the student model has a fraction $C$ of the teacher model's channels in its convolutional layers, with $C < 1$.

As discussed in Section \ref{kd-in-gans},  the challenge in applying knowledge distillation to GANs is the capacity mismatch between the teacher discriminator and the student generator. Our proposed solution to address this is the introduction of a second discriminator, the student discriminator $D_S$.
Similarly to the student generator $G_S$, we set the student discriminator $D_S$ to also have a fraction $C$ of the teacher discriminator's channels in its convolutional layers. Thus, the student models have approximately equal capacity, leading to more stable and effective adversarial training.

Overall, our proposed methodology, Student Discriminator Assisted Knowledge Distillation (SDAKD), consists of 3 sequential stages illustrated in Figure \ref{fig:main}. In the first stage, the teacher models are pre-trained (unless, of course, trained teacher models are readily available). Subsequently, the student generator and discriminator undergo supervised training from their teacher counterparts. In the final stage, adversarial training of the student networks is performed, combined with supervised training of the student generator.

\subsection{Feature Map Distillation}
In the latter two stages of SDAKD, both response-based supervision and Feature Map (FM) supervision are used. Among the numerous FM distillation approaches proposed in the literature, we adopt the MLP methodology \cite{liu2023simple}, because of its consistently high performance and low computational complexity. In the MLP methodology, the FM from the last layer of the student’s convolutional backbone is first transformed by a trainable two layer Multi Layer Perceptron (MLP). Subsequently, the L2 loss is computed between the transformed student FM and the corresponding teacher FM. We denote this loss as $\mathcal{L}_{MLP}$ in Figure \ref{fig:main}. In \cite{liu2023simple} the FM from the last layer of the student’s convolutional backbone had the same number of channels as the teacher`s FM. In our case, however, the student is constructed such that each layer has a fraction of the corresponding teacher’s channels. This modification in the student`s architecture is accommodated by adjusting the input dimensionality expected by the first layer of the MLP network.

\subsection{Training Strategy}
The SDAKD training strategy consists of three stages, as already introduced in Section \ref{over-appr}.

The first stage - if performed - encompasses pre-training the teacher networks for $n_1$ epochs, to replicate their original training procedure. The adversarial loss $\mathcal{L}_{{adv}_{T}}$, along with any additional losses utilized in the original work that introduced the teacher networks, are used.

In the second training stage, each of the randomly-initialized student networks is separately trained in a supervised way for $n_2$ epochs. This is motivated by the findings of \cite{li2020gan}, which showed that initiating adversarial training with randomly initialized weights is not optimal; it leads to lower image quality and possibly training instability. Specifically, in SDAKD the student generator is trained using both response-based and FM-based supervision, while the student discriminator is trained using only response-based supervision. Thus, the total loss for the student generator is defined as:

 \begin{equation}
\mathcal{L}_{G_{S}\_stage2} = \mathcal{L}_{sup} + \lambda_1 \mathcal{L}_{MLP}
\label{eq:stage2-lgs},
    \end{equation}

\noindent where $\mathcal{L}_{sup}$ is a supervised loss that compares the outputs of the student and teacher generators (\eg,  $\mathcal{L}_{1}$ loss,  $\mathcal{L}_{2}$ loss or Perceptual \cite{johnson2016perceptual} loss), $\mathcal{L}_{MLP}$ is the MLP FM distillation loss \cite{liu2023simple} and $\lambda_1$ is a loss weight coefficient. 
Additionally, the total loss for the student discriminator is defined as:

 \begin{equation}
\mathcal{L}_{D_{S}\_stage2} = \mathcal{L}_2(P_{real_S}, P_{real_T})
\label{eq:stage2-lds},
    \end{equation}

\noindent where $P_{real_S}$ and $P_{real_T}$ denote the probabilities that an input is classified as real by the student and teacher discriminators, respectively.

Finally, in the third stage, SDAKD performs adversarial training of the student networks, while continuing to also apply supervised training to the student generator. This final stage encompasses $n_3$ epochs, during which the total loss of the student generator is calculated as:

 \begin{equation}
\mathcal{L}_{G_{S}\_stage3} = \mathcal{L}_{sup} + \lambda_1 \mathcal{L}_{MLP} + \lambda_2 \mathcal{L}_{{adv}_{S}}
\label{eq:stage3-lgs},
    \end{equation}

\noindent where $\lambda_1$ and $\lambda_2$ are loss weight coefficients and $\mathcal{L}_{{adv}_{S}}$ is the adversarial loss of the student networks. The choice of this loss depends on the specific GAN model employed; it is not prescribed by SDAKD. Additionally, the loss for the student discriminator is defined as:

 \begin{equation}
\mathcal{L}_{D_{S}\_stage3} = \mathcal{L}_{{adv}_{S}}
\label{eq:stage3-lds}.
    \end{equation}

\noindent The adversarial losses $\mathcal{L}_{{adv}_{T}}$, $\mathcal{L}_{{adv}_{S}}$ may be implemented as instances of, \eg, the standard GAN \cite{goodfellow2014generative}, Wasserstein \cite{pmlr-v70-arjovsky17a} or Least Squares \cite{mao2017least} losses.

%% file: sec/4_experiments.tex
\section{Experiments}

\subsection{Models, Datasets and Metrics}

The proposed SDAKD distillation methodology is evaluated on two recent, well-performing and widely-adopted convolutional super-resolution GAN models, GCFSR \cite{he2022gcfsr} and Real-ESRGAN \cite{wang2021real}.

GCFSR \cite{he2022gcfsr} is a SOTA face super-resolution GAN model. Following the original work, we train GCFSR using the FFHQ dataset \cite{karras2019style}, which contains 70,000 high-quality $1024 \times 1024$ face images. For testing, we again adopt the same strategy as in the original paper and extract 100 random images from the CelebA-HQ validation dataset \cite{karras2017progressive}. Note that in the GAN literature it is often unclear whether test sets are used exclusively for testing or also for hyperparameter tuning and model selection. To ensure that the validation and test sets are disjoint, we construct and use a separate validation set by extracting an additional 100 random images from the CelebA-HQ validation dataset.

Real-ESRGAN \cite{wang2021real} is a widely-used and high-performing general image super-resolution model. Following
the original work, we train Real-ESRGAN on the combination of the DIV2K \cite{agustsson2017ntire}, Flickr2K \cite{timofte2017ntire} and OutdoorSceneTraining \cite{wang2018recovering} datasets. For validation and testing, we evenly divide the OutdoorSceneTraining test set, which contains 300 images in total.

Similarly to previous studies on GAN knowledge distillation \cite{hu2023discriminator, ren2021online, li2021revisiting}, we evaluate performance using the Fréchet Inception Distance (FID) \cite{heusel2017gans}.
FID measures the similarity between the distribution of real and generated samples. In the case of the GCFSR model, which supports upscaling factors in the range $[4, 64]$, we follow the original work and evaluate its performance at 16$\times$, 32$\times$, and 64$\times$ upscaling factors. In contrast, Real-ESRGAN supports upscaling of either 2$\times$ or 4$\times$. In our experiments, we evaluate the most challenging 4$\times$ variant.

\begin{table*}[t]
\centering
\small 
\resizebox{\textwidth}{!}{ 
\begin{tabular}{l|cc|ccc|ccc}
\toprule
Method  & \makecell{$\times$ Student\\Generator\\Channels} 
        & \makecell{$\times$ Discr-\\iminator\\Channels} 
        & \makecell{Validation\\FID 64$\times$ ↓} 
        & \makecell{Validation\\FID 32$\times$ ↓} 
        & \makecell{Validation\\FID 16$\times$ ↓}
        & \makecell{Test\\FID 64$\times$ ↓} 
        & \makecell{Test\\FID 32$\times$ ↓} 
        & \makecell{Test\\FID 16$\times$ ↓} \\
\midrule
GCFSR (original) \cite{he2022gcfsr} & 1 & 1 & \textcolor{gray}{57.15} & \textcolor{gray}{43.34} & \textcolor{gray}{30.48} & - & - & -\\
GCFSR (reproduced)  & 1 & 1 & 57.01 & 43.87 & 30.95 & 56.71 & 45.72 & 31.54 \\
\midrule
GCFSR \cite{he2022gcfsr} - (1/2, 1)   & 1/2 & 1 & 58.59 & 47.04 & 33.12 & 58.14 & 49.32 & 33.82 \\
GCFSR \cite{he2022gcfsr}  - (1/2, 1/2) & 1/2 & 1/2 & 58.51 & 46.71 & 32.96 & 58.03 & 48.52 & 33.25 \\
OMGD \cite{ren2021online}       & 1/2 & 1 & 105.93 & 91.52 & 65.14 & 110.46 & 94.89 & 68.52\\
DCD \cite{hu2023discriminator}  & 1/2 & 1 & 91.18 & 72.79 & 52.34 & 96.86 & 76.66 & 55.68\\
\makecell[l]{ DCD with modified \\ adversarial training}      & 1/2 & 1 & 67.32 & 53.95 & 38.60 & 67.04 & 55.87 & 39.91\\
SDAKD (ours)  &  1/2 & 1/2 & \textbf{57.48} & \textbf{44.91} & \textbf{31.67} & \textbf{57.94} & \textbf{46.38} & \textbf{33.15} \\
\midrule
GCFSR \cite{he2022gcfsr}  - (1/4, 1)  & 1/4 & 1 & 67.30 & 55.41 & 41.28 & 67.12 & 56.32 & 41.68 \\
GCFSR \cite{he2022gcfsr} - (1/4, 1/4)  & 1/4 & 1/4 & 64.12 & 51.08 & 37.82 & 64.22 & 53.91 & 38.65 \\
SDAKD (ours)  & 1/4 & 1/4 & \textbf{60.22} & \textbf{48.15} & \textbf{33.81} & \textbf{58.61} & \textbf{49.20} & \textbf{34.36} \\
\midrule
GCFSR \cite{he2022gcfsr} - (1/8, 1)   & 1/8 & 1 & 74.85 & 61.56 & 45.82 & 74.42 & 63.16 & 46.47 \\
GCFSR \cite{he2022gcfsr} - (1/8, 1/8) & 1/8 & 1/8 & 71.42 & 58.11  & 42.79 & 71.36 & 60.53 &  43.42\\
SDAKD (ours)  & 1/8 & 1/8 & \textbf{67.74} & \textbf{54.42} & \textbf{39.81} & \textbf{66.76} & \textbf{55.04} & \textbf{40.77} \\
\bottomrule
\end{tabular}}
\caption{Comparison of the proposed SDAKD methodology with existing methods on the GCFSR model. Our validation set differs from that of the original GCFSR study, as they selected a different set of 100 random images from the CelebA-HQ dataset. Hence, the results in gray (first row of this table) are not directly comparable with the rest.}
\label{table:gcfsr-results}
\end{table*}

\begin{table*}[]
\centering
\small 
\begin{tabular}{l|cc|c|c}
\toprule
Method  & $\times$ Student Gener. Channels 
        & $\times$ Discrim. Channels 
        & Valid. FID 4$\times$ ↓ 
        & Test FID 4$\times$ ↓ \\
\midrule
Real-ESRGAN \cite{wang2021real} & 1 & 1 &  44.89 & 45.62   \\
\midrule
Real-ESRGAN \cite{wang2021real} - (1/2, 1)  & 1/2 & 1 & 58.46 &  58.72\\
Real-ESRGAN \cite{wang2021real} - (1/2, 1/2) & 1/2 & 1/2 & 57.90 & 58.24 \\
OMGD \cite{ren2021online}       & 1/2 & 1 &  54.47 & 55.89 \\
DCD \cite{hu2023discriminator}  & 1/2 & 1 & 52.90 & 53.43 \\
\makecell[l]{DCD with modified adversarial training}      
                             & 1/2 & 1 &  54.62 & 54.93 \\
SDAKD (ours)  &  1/2 & 1/2 & \textbf{49.08} & \textbf{49.57}\\
\midrule
Real-ESRGAN \cite{wang2021real}  - (1/4, 1) & 1/4 & 1 & 63.65 & 63.91 \\
Real-ESRGAN \cite{wang2021real} - (1/4, 1/4) & 1/4 & 1/4 & 62.51 & 62.94\\
SDAKD (ours)  & 1/4 & 1/4 & \textbf{49.93} & \textbf{50.49}   \\
\midrule
Real-ESRGAN \cite{wang2021real}  - (1/8, 1) & 1/8 & 1 & 74.72 & 75.07\\
Real-ESRGAN \cite{wang2021real} - (1/8, 1/8) & 1/8 & 1/8 & 71.38 & 71.74   \\
SDAKD (ours)  & 1/8 & 1/8 & \textbf{55.82} &  \textbf{56.47} \\
\bottomrule
\end{tabular}
\caption{Comparison of the proposed SDAKD methodology with existing methods on the Real-ESRGAN model.}
\label{table:realesrgan-results}
\end{table*}

\begin{table}[]
\centering
\small
\begin{tabular}{ c|c|c}
 \toprule
 \makecell{$\times$ Number of\\ Channels} & \makecell{GCFSR\\ Speed-up} & \makecell{Real-ESRGAN\\ Speed-up} \\
 \midrule
 1  & none (1.00$\times$) & none (1.00$\times$) \\
1/2 & 1.98$\times$ & 1.77$\times$ \\
1/4 & 3.43$\times$ & 1.96$\times$ \\
1/8 & 4.35$\times$ & 2.08$\times$ \\
 \bottomrule
\end{tabular}
\caption{Inference time speed-up based on the number of channels of the GCFSR and Real-ESRGAN generator, following the application of the proposed SDAKD distillation method.}
\label{table:speed-up}
\end{table}

\subsection{Implementation Details}
In our experiments, each SDAKD training stage was set to encompass the same number of epochs, \ie $n_1 = n_2 = n_3 = n$. In the case of the GCFSR network, the learning curve reported in the original study illustrates its fast convergence. Indeed, in our experiments we were able to reproduce the results reported in \cite{he2022gcfsr} with 30 training epochs (see Table \ref{table:gcfsr-results}), rather than the full 100. Therefore, in our methodology we set $n=30$ for the GCFSR network. In the case of the Real-ESRGAN network, we followed the original work and trained for 1.4M steps, which correspond to $n=300$ epochs.

With respect to loss functions, we employed in our experiments the ones used in the original works \cite{ren2021online, hu2023discriminator} of each examined GAN. Specifically, we set the supervised loss $\mathcal{L}_{sup}$ as a combination of the $\mathcal{L}_{1}$ and Perceptual \cite{johnson2016perceptual} losses. 
For the adversarial losses $\mathcal{L}_{{adv}_{T}}$ and $\mathcal{L}_{{adv}_{S}}$, we used the standard GAN loss \cite{goodfellow2014generative} in the case of Real-ESRGAN. For GCFSR, we used the non-saturating logistic loss, which is an identical formulation of the standard GAN loss, but more numerically stable. Additionally, for the loss weight coefficient $\lambda_1$, we used the values used in the original works, while for $\lambda_2$, we set its value empirically such that all three components of Equation \ref{eq:stage3-lgs} are approximately equal at the start of the training. Specifically, we set $\lambda_1 = 1$ and $\lambda_2 = 0.1$.

Concerning FM distillation in SDAKD, we initially experimented with applying it to both the student generator and student discriminator. However, these early experiments did not show a positive effect on the discriminator performance. This may be attributed to architectural differences between the discriminator and the networks where the MLP FM distillation method has previously shown strong performance. Specifically, MLP has been used so far in network architectures where a convolutional backbone is followed by a considerable number of fully connected layers. In contrast, the discriminators in our experiments consist of only a single layer beyond the convolutional backbone. This architectural difference limits the effectiveness of the MLP FM-based supervision in the discriminator, since MLP supervises the backbone`s last convolutional layer FM. Thus, in the reported experiments we apply FM-based knowledge distillation only to the student generator.

Throughout training, we evaluate the student generator’s performance after each epoch. Model selection is performed using the validation set (according to the best validation FID score). For GCFSR, where we have multiple upscaling factors, the average FID across the 16$\times$, 32$\times$ and 64$\times$ upscaling factors on the validation set is considered.

Concerning Stage 1 of SDAKD, this is needed when distilling GCFSR, because for this network only the pre-trained generator is publicly available. Therefore, we carried out this training stage to pre-train both the teacher generator and discriminator. In contrast, for Real-ESRGAN both the pre-trained generator and discriminator are publicly available, thus Stage 1 of SDAKD was omitted.

\subsection{Results and Comparisons}
We compare our proposed SDAKD methodology to the SOTA GAN knowledge distillation methods DCD \cite{hu2023discriminator} and OMGD \cite{ren2021online}. Additionally, we compare against two naive baselines in which knowledge distillation is not performed; instead, the number of channels is reduced in either only the generator or in both the generator and the discriminator, and the compacted GAN is trained from scratch. 

\begin{figure}[]
    \centering
    \includegraphics[width=0.8\linewidth]{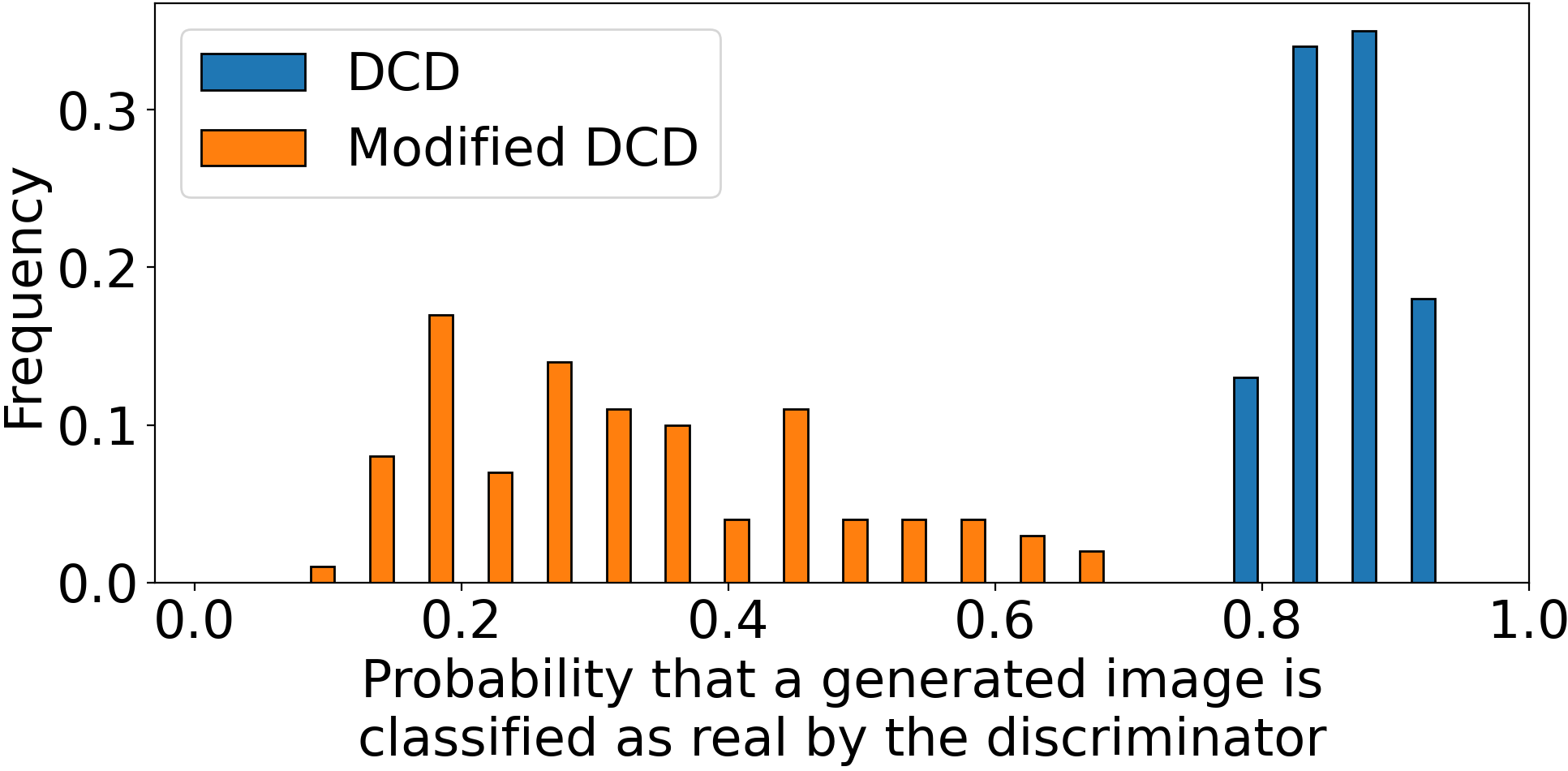}
    \caption{Comparison of the DCD versus modified DCD output distribution of the discriminator that is given as input an image generated by the student generator.}
    \label{fig:dcd-distribution}
\end{figure}

The experimental results of applying SDAKD and the competing distillation methods to the GCFSR network are reported in Table \ref{table:gcfsr-results}. SDAKD consistently outperforms the baselines across all three evaluated channel sizes. 
In contrast, DCD and OMGD perform poorly in this experiment, being greatly outperformed by the baselines in the least-challenging $C=1/2$ setting (for this, there was no point in further experimenting with them in the more challenging $C=1/4$, $C=1/8$ settings). DCD and OMGD have been successfully applied to Pix2Pix \cite{isola2017image} and StyleGAN \cite{karras2019style}, which are earlier super-resolution networks with an output image resolution of 256×256, whereas in our setting the output resolution of GCFSR is substantially higher at 1024×1024. We hypothesize that the higher target resolution results in a significantly larger sample space, which necessitates the use of a discriminator to achieve good generalization, instead of only relying on supervised training as OMGD does. Furthermore, we hypothesize that when the resolution (and consequently the sample space) is small, the outputs of both the teacher and student generator are sufficiently similar; this renders the teacher discriminator in DCD able to effectively supervise adversarially both generators, while only being trained against the teacher generator. Contrarily, a larger sample space results in the outputs of the teacher and student generator being less similar, and this allows the student generator to successfully fool a discriminator that is not trained against (also) this generator. To test these hypotheses, we conducted an experiment with a modified DCD in which the teacher discriminator was trained adversarially against both generator networks (denoted ``DCD with modified adversarial training'' in Table \ref{table:gcfsr-results}). The results show that although SDAKD still outperforms this modified DCD, the performance gap is notably smaller. The output distributions illustrated in Figure \ref{fig:dcd-distribution} clearly show how the student generator can successfully fool the discriminator (output skewed towards 1) in the original DCD, thereby preventing the latter from providing meaningful training signals to the student generator; this is not the case in the modified DCD method. Finally, the fact that SDAKD outperforms the naive baseline (in which both the generator and discriminator feature a similarly reduced number of channels) indicates that balancing network capacity alone is insufficient; a well-performing teacher is also required for optimal student generator training.

Table \ref{table:realesrgan-results} presents the results of applying SDAKD to the Real-ESRGAN network. As before, we compare our approach to the two naive baselines as well as OMGD \cite{ren2021online} and DCD \cite{hu2023discriminator}. We observe that SDAKD consistently outperforms all competing approaches across all evaluated channel sizes $C$. The performance gap between SDAKD and the OMGD and DCD methods is smaller, compared to the experiments with the GCFSR network. This observation is consistent with our hypothesis regarding image resolution, as the output resolution of Real-ESRGAN during training is a more modest 400×400 pixels. Additionally, we again assessed the ``DCD with modified adversarial training'', specified above. In contrast to the results in Table \ref{table:gcfsr-results}, for Real-ESRGAN the standard DCD methodology slightly outperforms the modified one. As already discussed, we attribute this to the teacher discriminator being able to provide effective adversarial supervision to the student generator at lower resolutions, even without being trained against it. Moreover, training the discriminator adversarially against both generators compromises the training of the teacher generator, as the discriminator must balance between two competing tasks. 

Finally, Table \ref{table:speed-up} reports how the number of channels of the GCFSR and Real-ESRGAN generator affects inference time. Inference times were measured by averaging over 50 forward passes. Both networks achieved substantial speed-up by reducing the number of channels to half. For GCFSR, further reducing the channels to 1/4 yielded additional gains, although diminishing returns appeared at 1/8. In contrast, Real-ESRGAN exhibited diminishing returns in inference speed already at 1/4 of the original channels. In general, reducing the number of channels in both networks resulted in significant inference acceleration. However, the point at which diminishing returns appear is highly dependent on the network architecture.

\begin{figure}[]
\begin{center}
\begin{tabular}{c}
\includegraphics[width=0.8\linewidth]{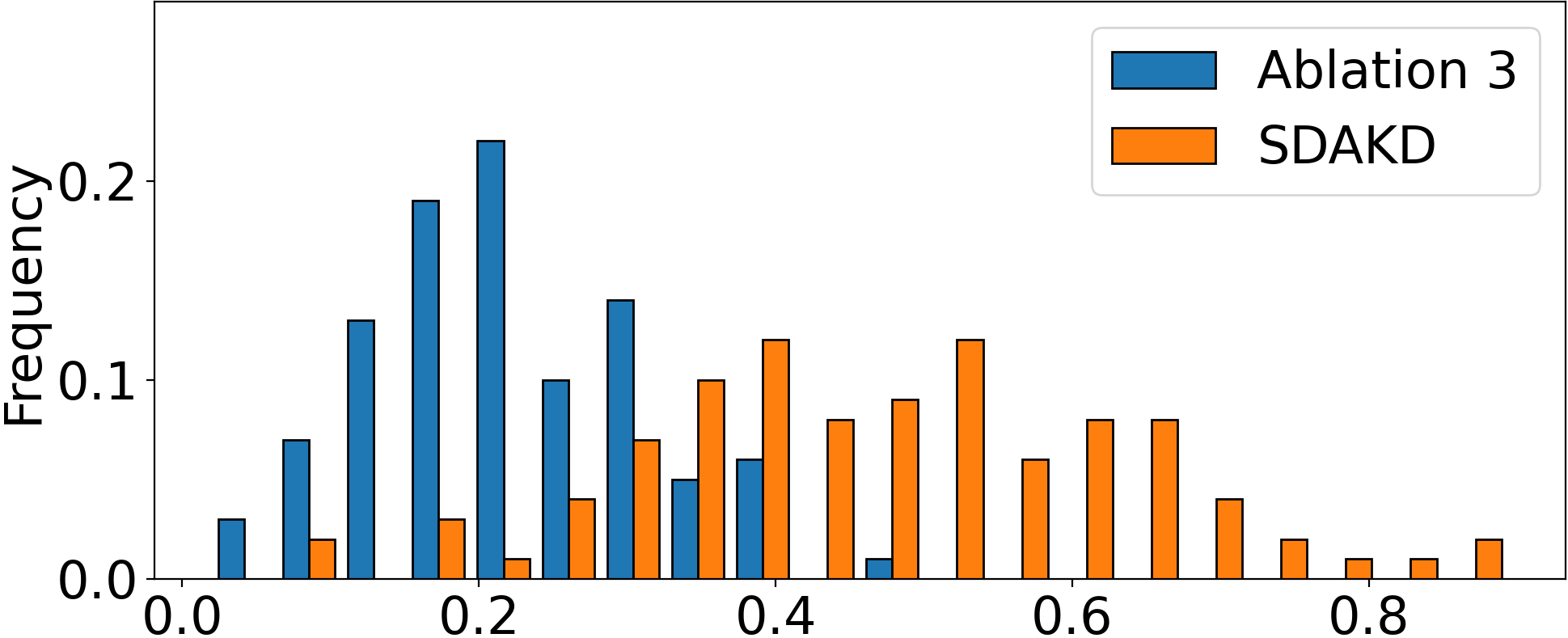}\\
 \includegraphics[width=0.8\linewidth]{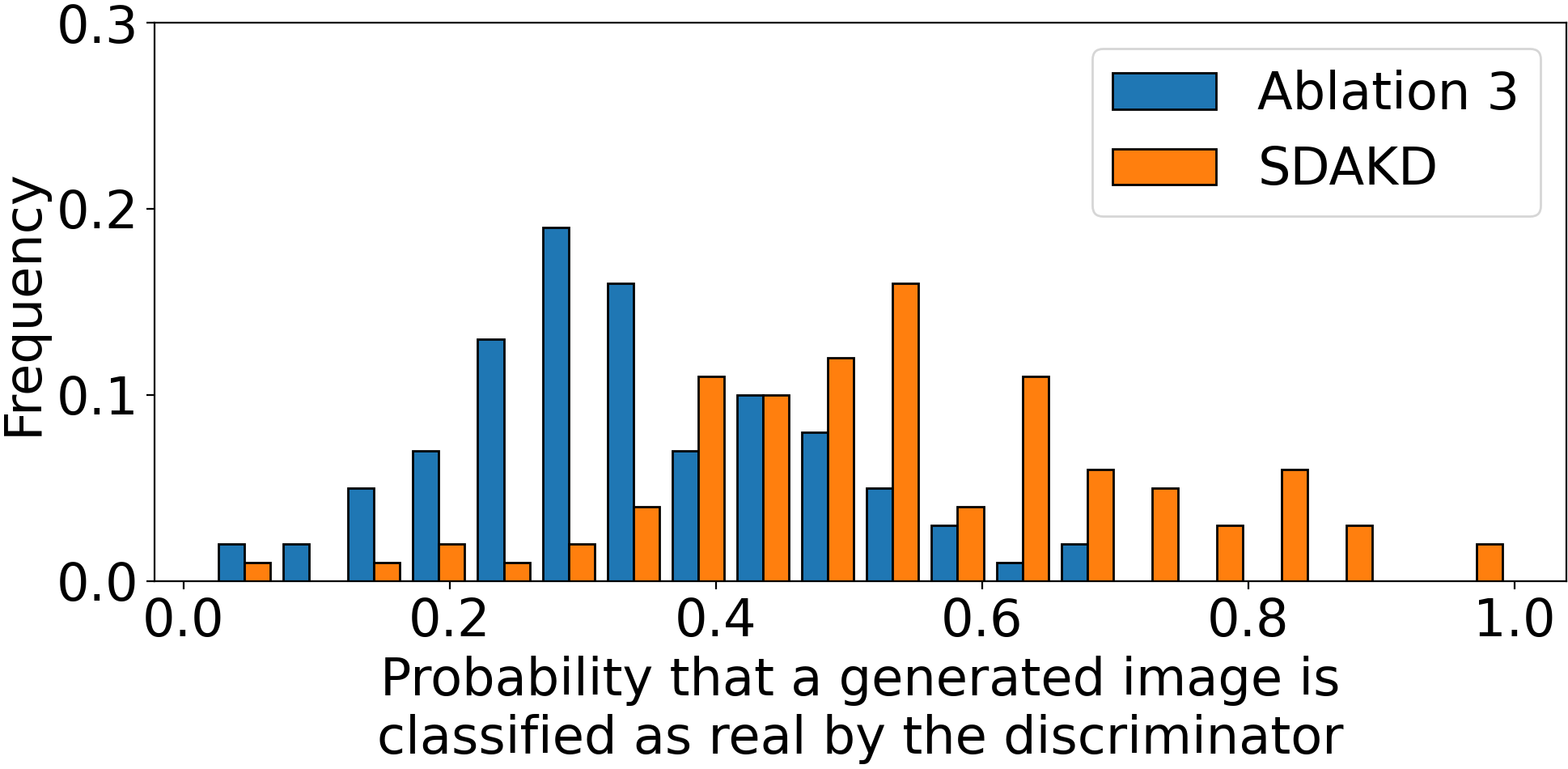}  
\end{tabular}
\end{center}
\caption{Output distributions of the discriminator given a generated image as input, using the pre-trained teacher versus the student discriminator, for GCFSR (top) / Real-ESRGAN (bottom).}
\label{fig:discriminator-distributions}
\end{figure}

\begin{table*}[]
\centering
\small 
\begin{tabular}{l|ccc|ccc|ccc}
\toprule
Method  & \makecell{\# Super-\\vised-only \\epochs ($n_2$)} 
        & \makecell{FM\\ Distil-\\lation} 
        & \makecell{Student\\ Discr-\\iminator}  
        & \makecell{Validation\\FID 64$\times$ ↓} 
        & \makecell{Validation\\ FID 32$\times$ ↓} 
        & \makecell{Validation \\ FID 16$\times$ ↓} 
        & \makecell{Test FID \\ 64$\times$ ↓} 
        & \makecell{Test FID \\ 32$\times$ ↓} 
        & \makecell{Test FID \\ 16$\times$ ↓}\\
        
\midrule
GCFSR (rep.) &&&&57.01&43.87&30.95&56.71&45.72&31.54\\
\midrule
Ablation 1    & 0 &&&61.57&50.32&35.26&60.03&49.78&36.56\\ 
Ablation 2  & 30 & & & 61.37&49.78&35.12&59.69&49.50&36.34\\
Ablation 3 & 30 & \checkmark && 60.89&49.08&34.75&59.42&49.46&35.92\\
SDAKD (ours)   & 30 & \checkmark & \checkmark & \textbf{60.22} & \textbf{48.15} & \textbf{33.81} & \textbf{58.61} & \textbf{49.20} & \textbf{34.36}\\
\bottomrule
\end{tabular}
\caption{Ablations for the GCFSR network, with the student having one quarter the number of the original channels.}
\label{table:ablation-results}
\end{table*}

\begin{figure*}[]
\setlength{\tabcolsep}{1pt}
\renewcommand{\arraystretch}{0.4}
\begin{center}
\begin{tabular}{cccccc}
Input & Ground Truth & GCFSR & GCFSR 1/2 & GCFSR 1/4 & GCFSR 1/8 \\
\includegraphics[width=0.1625\linewidth]{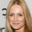}&
\includegraphics[width=0.1625\linewidth]{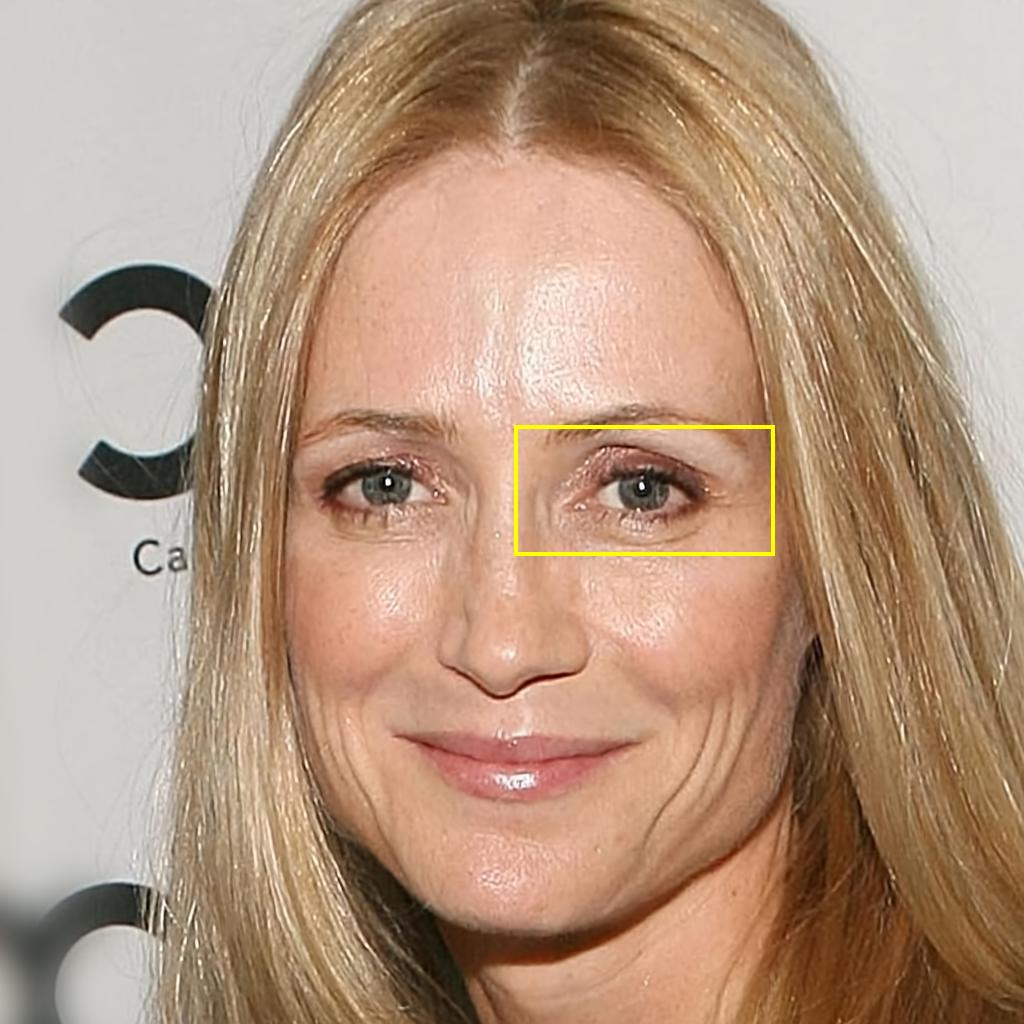}&
\includegraphics[width=0.1625\linewidth]{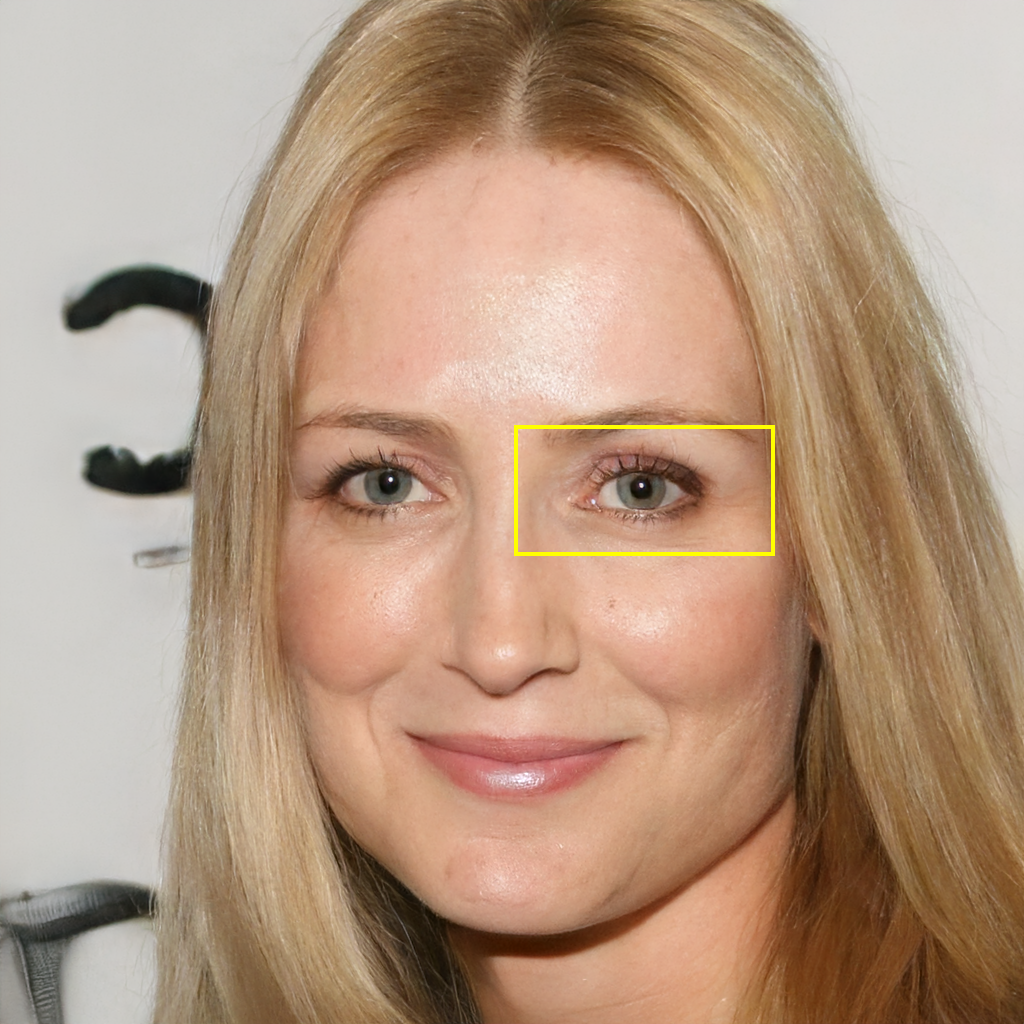}&
\includegraphics[width=0.1625\linewidth]{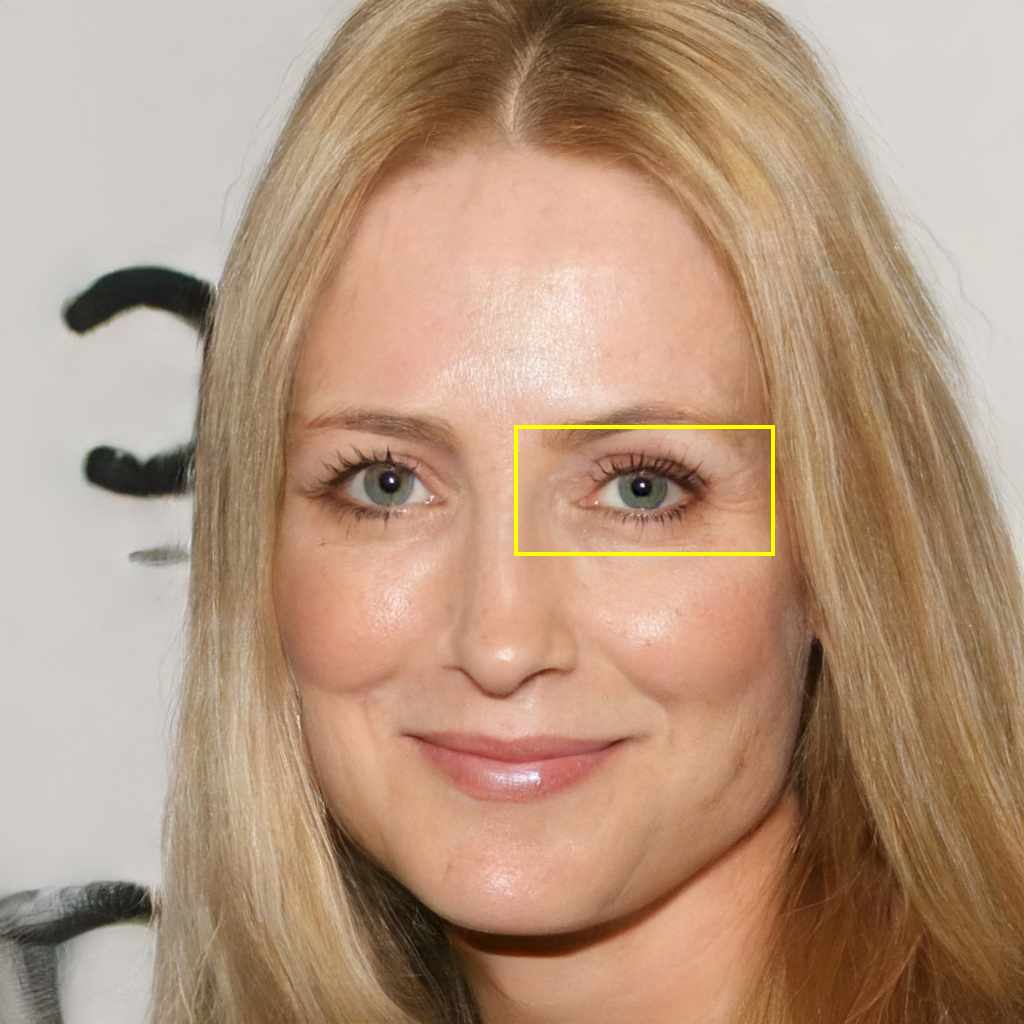}&
\includegraphics[width=0.1625\linewidth]{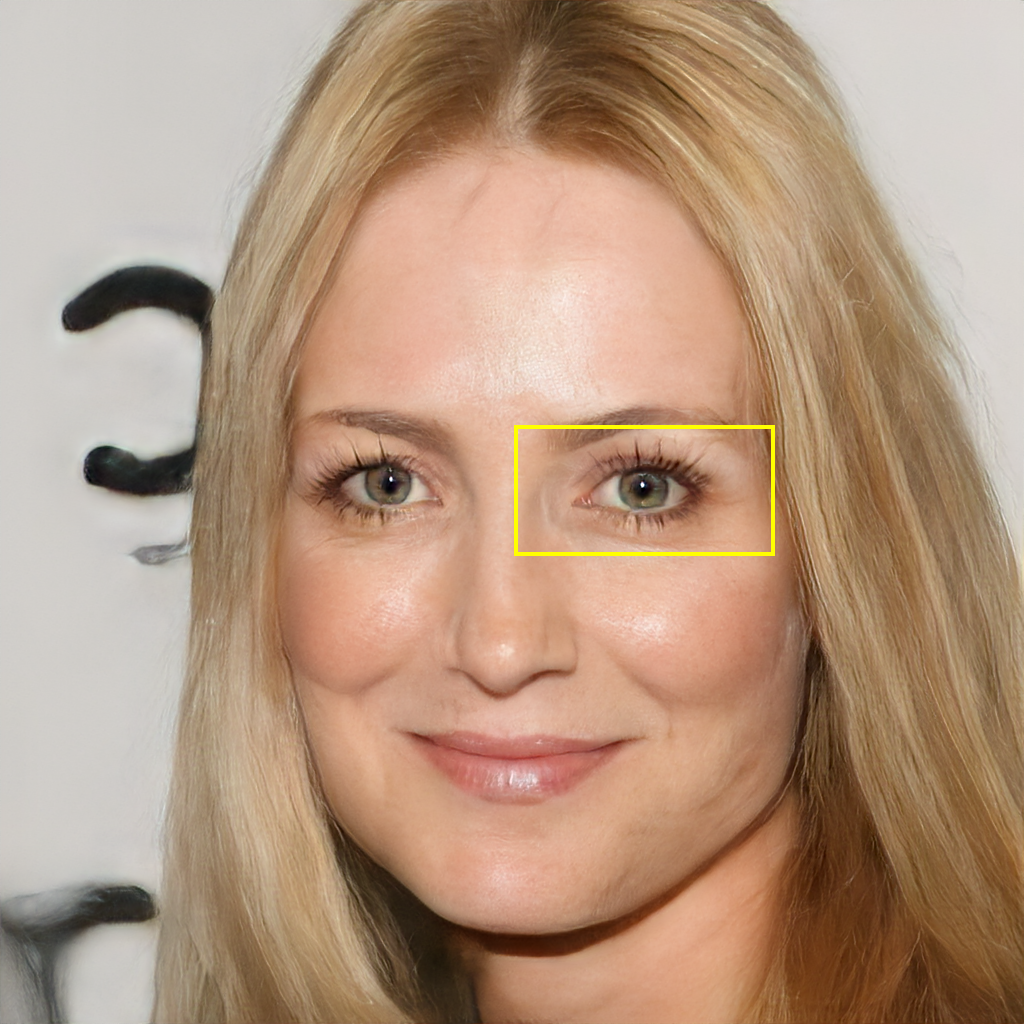}&
\includegraphics[width=0.1625\linewidth]{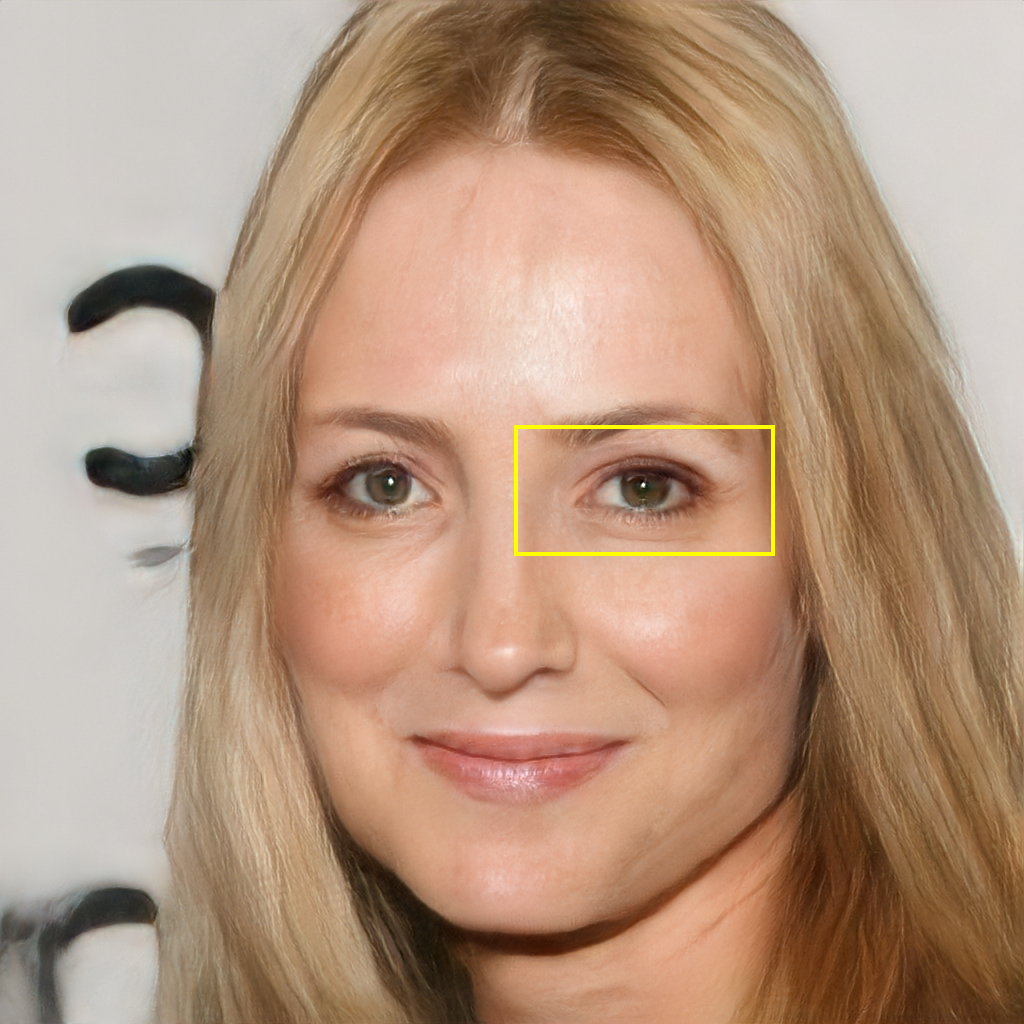} \\
&
\includegraphics[width=0.1625\linewidth]{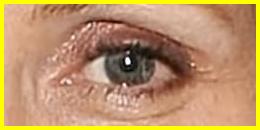}&
\includegraphics[width=0.1625\linewidth]{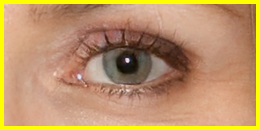}&
\includegraphics[width=0.1625\linewidth]{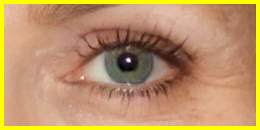}&
\includegraphics[width=0.1625\linewidth]{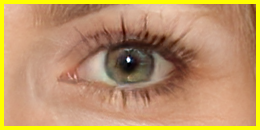}&
\includegraphics[width=0.1625\linewidth]{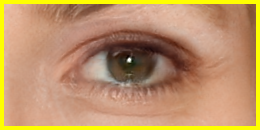} \\ 
\includegraphics[width=0.1625\linewidth]{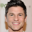}&
\includegraphics[width=0.1625\linewidth]{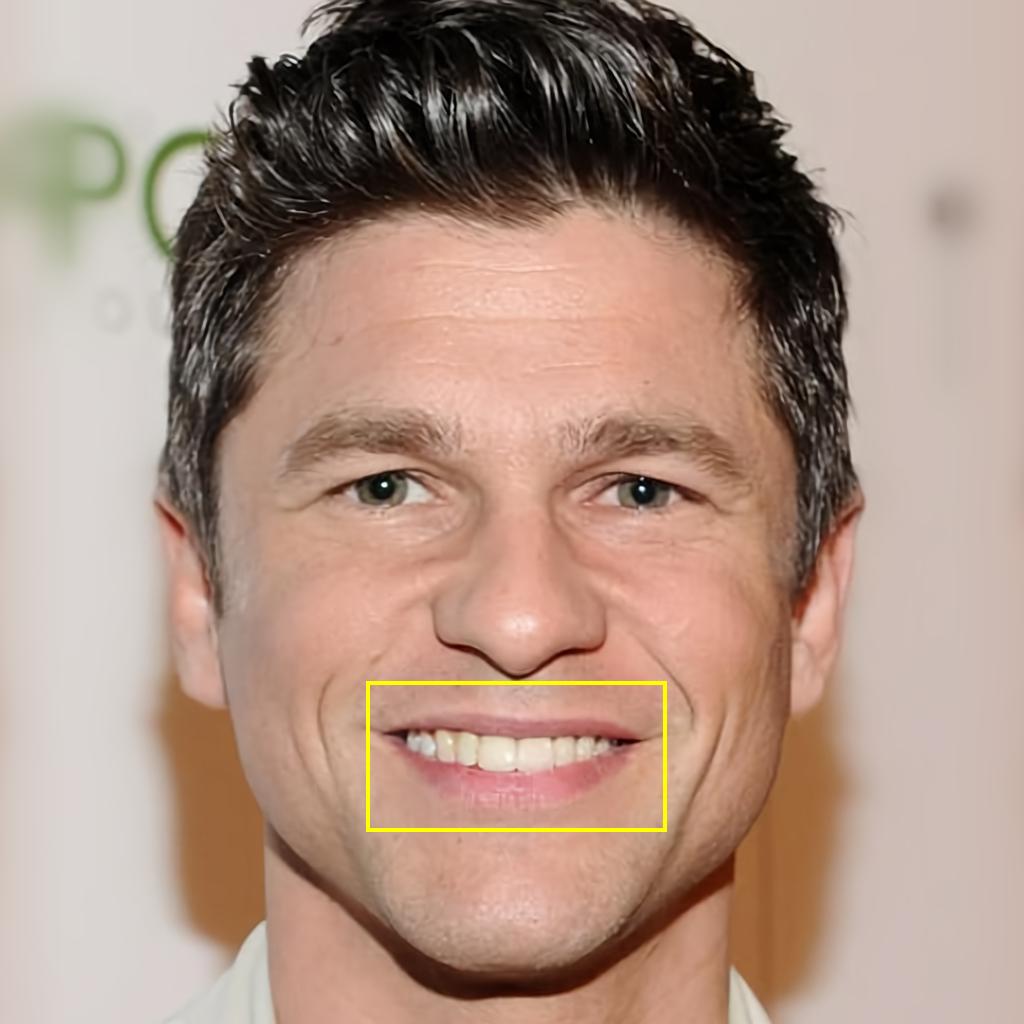}&
\includegraphics[width=0.1625\linewidth]{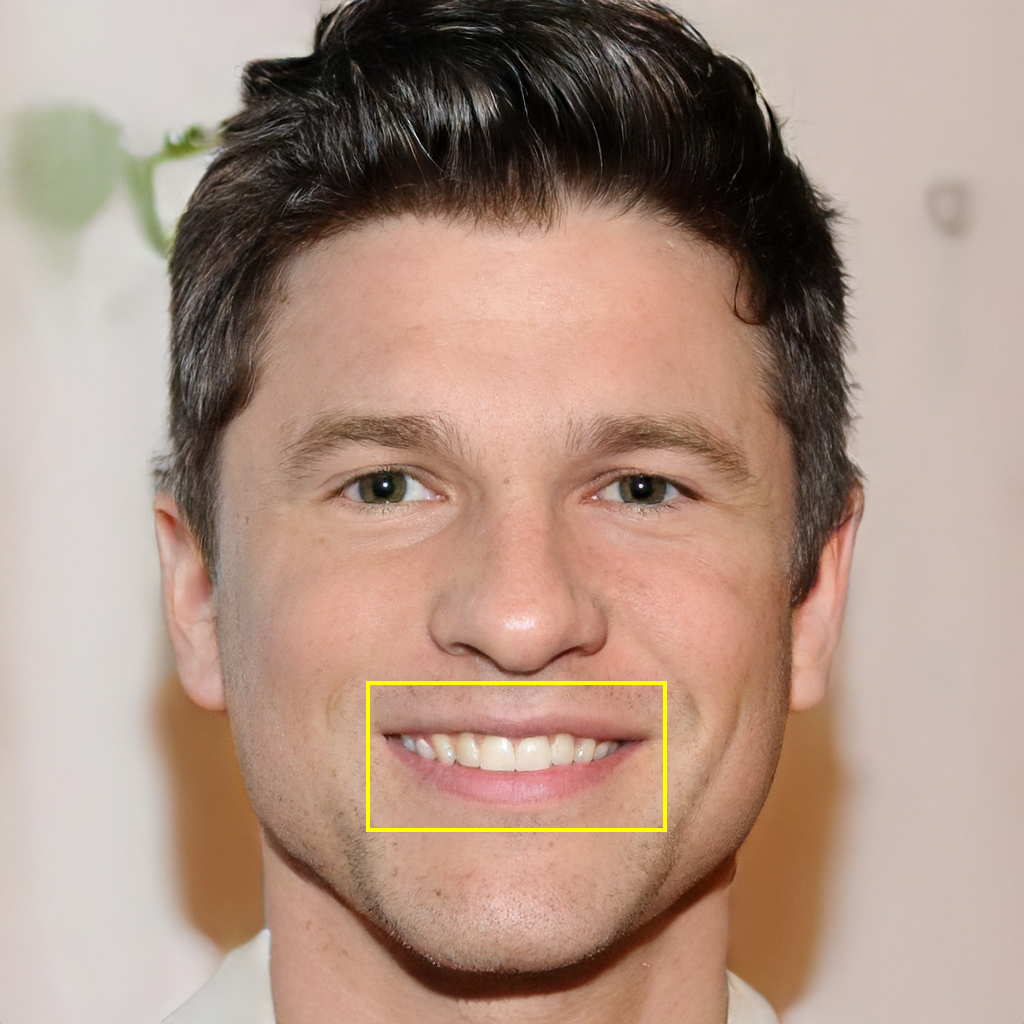}&
\includegraphics[width=0.1625\linewidth]{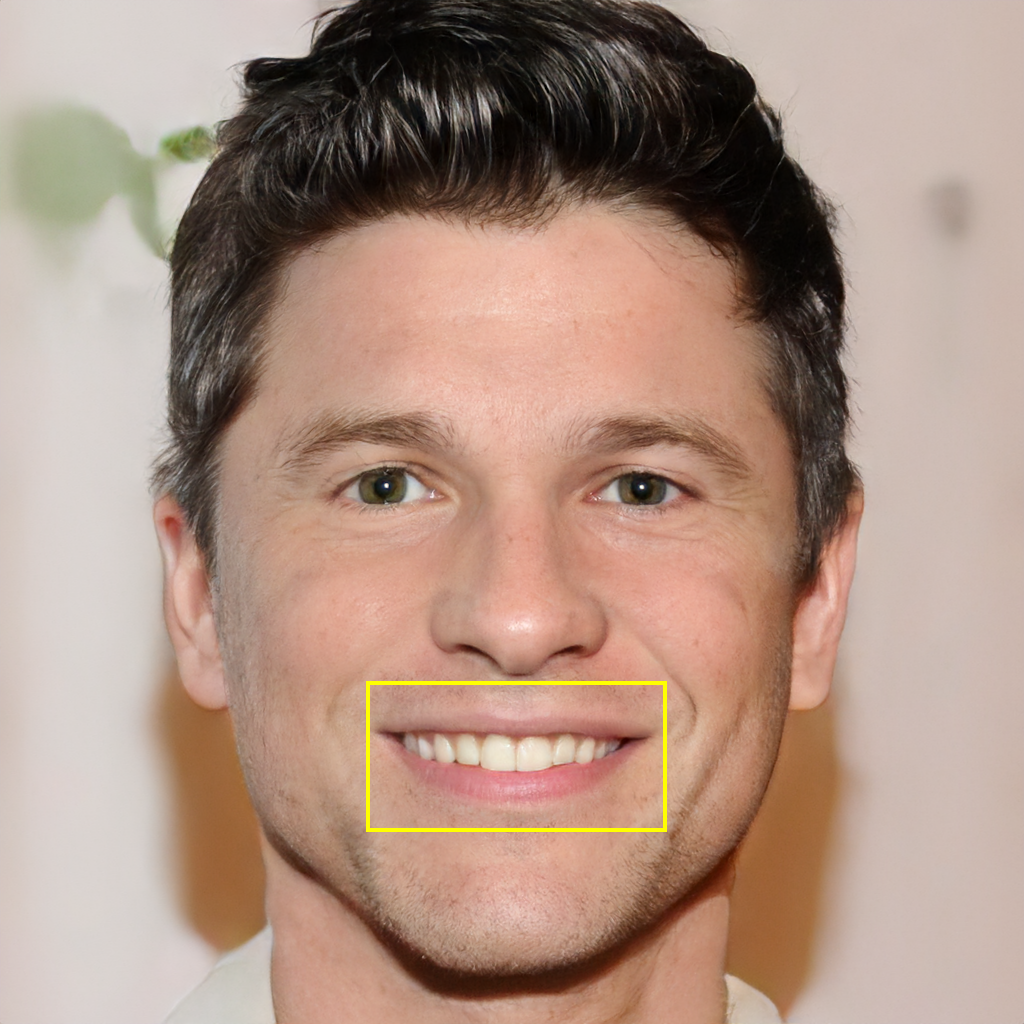}&
\includegraphics[width=0.1625\linewidth]{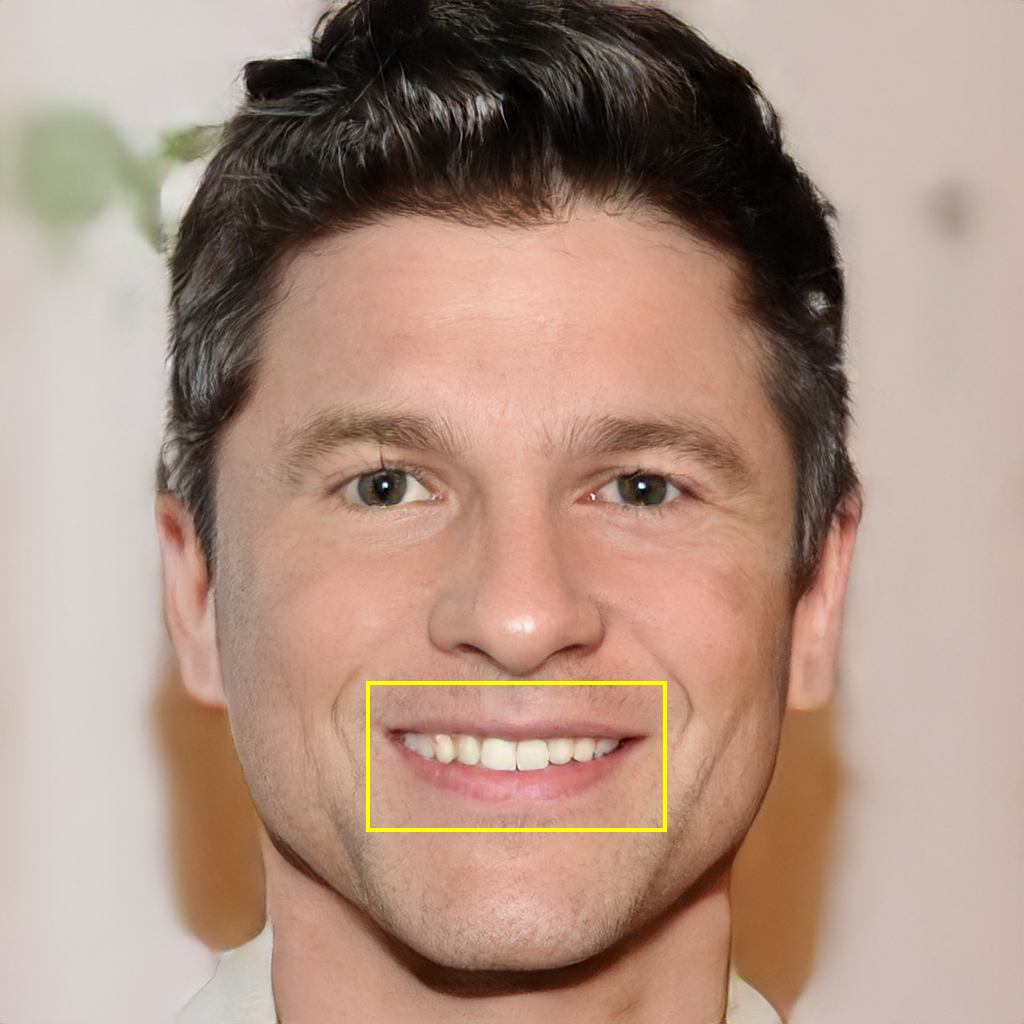}&
\includegraphics[width=0.1625\linewidth]{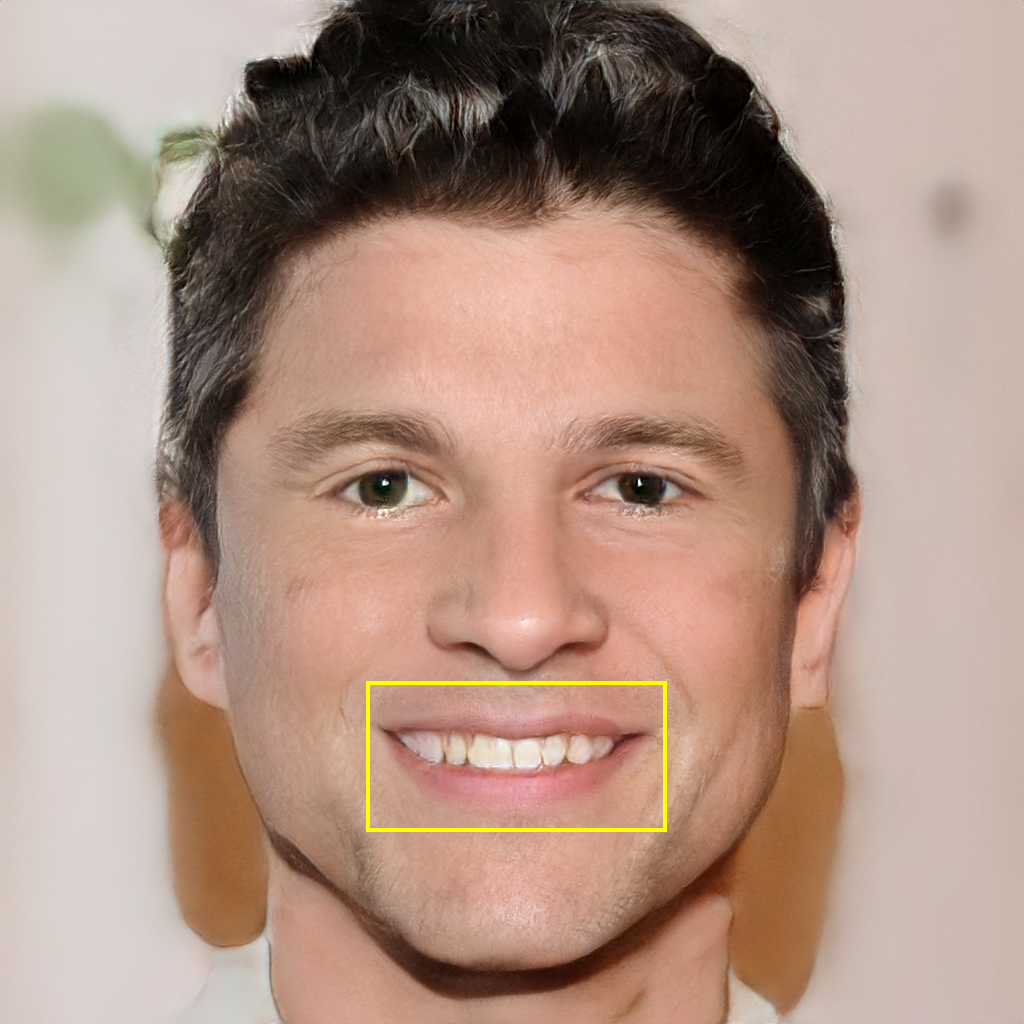} \\
&
\includegraphics[width=0.1625\linewidth]{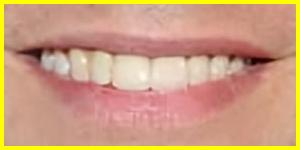}&
\includegraphics[width=0.1625\linewidth]{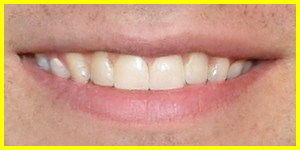}&
\includegraphics[width=0.1625\linewidth]{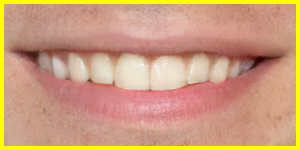}&
\includegraphics[width=0.1625\linewidth]{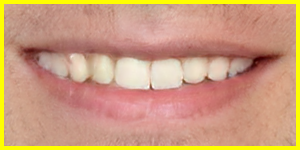}&
\includegraphics[width=0.1625\linewidth]{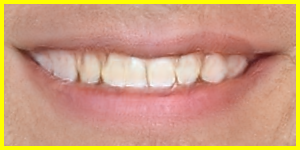} \\
\end{tabular}
\end{center}
\caption{Qualitative results of 32$\times$ upsampling using our proposed SDAKD methodology on the GCFSR network. The SDAKD student models have 1/2, 1/4 and 1/8 the number of channels of the original generator network. Results are compared against the ground truth and the original GCFSR model, which was also used as teacher for the knowledge distillation. Samples taken from our test set, derived from the CelebA-HQ dataset. The input image size is 32×32 pixels, while the output size is 1024×1024 pixels.}
\label{fig:qualitative}
\end{figure*}

\subsection{Ablations}
To assess the performance impact of the three main elements of SDAKD (\ie the newly-introduced student discriminator, the supervised-only training stage (Stage 2), and the MLP FM-based distillation), we conducted an ablation study on the GCFSR network using a student generator with $C=1/4$ the number of channels of the original model. The results are presented in Table \ref{table:ablation-results}. We observe that, while all three of the above elements had an positive affect on performance, the utilization of a student discriminator (instead of the teacher discriminator) for the adversarial training of the student generator had the largest effect.

To assess whether the capacity mismatch between the student generator and the teacher discriminator is alleviated by employing a student discriminator, we analyzed the discriminator output distributions when given as input an image generated by the student generator. Specifically, we compared the discriminator output distribution of SDAKD with that of its variant that utilizes the teacher discriminator instead of the student one (this variant corresponds to ``Ablation 3'' in Table \ref{table:ablation-results}). To estimate the discriminator`s output distribution, we first passed 100 training images through the student generator to obtain generated samples, which were then fed into the discriminator and its outputs were recorded. The results are presented in Figure \ref{fig:discriminator-distributions}. In these experiments, the student generator has 1/4 the number of channels of the teacher. Theoretically, in a perfectly trained GAN, the generator would have learned the true data distribution, and thus the generated samples would be indistinguishable from the real samples. Hence, the output of the discriminator would always be a random guess (\ie probability of 50\%) \cite{goodfellow2014generative}. In practice, however, the networks do not reach perfect equilibrium and the discriminator`s true output distribution deviates from the theoretical one. For instance, if the discriminator is too strong or too weak, the discriminator output values are skewed towards 0 or 1, respectively \cite{pmlr-v70-arjovsky17a}. In our experiments with both GCFSR and Real-ESRGAN, when using the pre-trained teacher discriminator the output distribution is closer to 0, indicating that, as expected, the teacher discriminator is stronger than the student generator. In contrast, when using the student discriminator, the output distribution is shifted to the right, centered approximately at 0.5. This is an indication that the utilization of the student discriminator makes the student adversarial training more balanced. Additionally, by examining both Figures \ref{fig:dcd-distribution} and \ref{fig:discriminator-distributions}, we observe that the output distributions of Ablation 3 and the modified DCD methodology are very similar, as they both utilize the teacher discriminator for the adversarial training of the student generator.

\subsection{Qualitative Results}
Qualitative results of 32$\times$ upscaling with SDAKD-distilled versions of GCFSR are shown in Fig. \ref{fig:qualitative}. To highlight fine-grained details, we also provide zoomed-in views of key facial features, such as the eyes and mouth. These examples indicate that the GCFSR`s performance does not significantly decline when applying SDAKD for $C=1/2$ or $C=1/4$ of its original number of channels. At the more aggressive $C=1/8$ rate, the generated images remain visually pleasing (compared to the input ones), but there is a clearly perceptible decline in quality in facial areas of fine-grained detail, compared to the $C=1/4$ setting. 

%% file: sec/5_conclusions.tex
\section{Conclusion}
We presented SDAKD, a new knowledge distillation methodology for GANs. The contributions of SDAKD are the introduction of a second (student) discriminator, the inclusion of a supervised-only training stage for both student networks, and the integration of a modified MLP FM distillation approach. We evaluated SDAKD on two well-performing super-resolution GANs, GCFSR and Real-ESRGAN. SDAKD was experimentally shown to outperform the SOTA methods for GAN knowledge distillation. Detailed analysis and discussion of the shortcomings of the latter methods, as well as ablations of the proposed SDAKD, provided insight on why the latter performs well and how the challenges of performing knowledge distillation on GANs can be overcome.